%% file: iclr2026_conference.tex
\documentclass{article} % For LaTeX2e
\usepackage{iclr2026_conference,times}

\input{math_commands.tex}

\usepackage{hyperref}
\usepackage{url}
\usepackage{booktabs}       % professional-quality tables
\usepackage{amsfonts}       % blackboard math symbols
\usepackage{nicefrac}       % compact symbols for 1/2, etc.
\usepackage{microtype}      % microtypography
\usepackage{xcolor}         % colors
\usepackage{graphicx} 
\usepackage{multirow}
\usepackage{diagbox}
\usepackage{amsmath}
\usepackage{amssymb}
\usepackage{bbding}
\usepackage{makecell}
\usepackage{colortbl}
\usepackage{enumitem}
\usepackage{wrapfig}
\definecolor{highlight_blue}{RGB}{0, 255, 255}
\definecolor{highlight_green}{RGB}{0, 255, 0}
\definecolor{green_color}{RGB}{78, 167, 46}
\definecolor{blue_color}{RGB}{34, 131, 200}

\title{Chat-CBM:\\Towards Interactive Concept Bottleneck Models with Frozen Large Language Models}

\author{
\begin{tabular}{c}
Hangzhou He\textsuperscript{1,2,3} \quad
Lei Zhu\textsuperscript{1,2,3} \quad
Kaiwen Li\textsuperscript{1,2,3} \quad
Xinliang Zhang\textsuperscript{1,2,3} \quad
Jiakui Hu\textsuperscript{1,2,3}\vspace{0.1cm}\\
Ourui Fu\textsuperscript{1,2,3} \quad
Zhengjian Yao\textsuperscript{1,2,3} \quad
Yanye Lu\textsuperscript{1,2,3,}\thanks{Corresponding author: yanye.lu@pku.edu.cn}\vspace{0.1cm}
\end{tabular} \\
\textsuperscript{1}Department of Biomedical Engineering, College of Future Technology, Peking University \\
\textsuperscript{2}Institute of Medical Technology, Peking University Health Science Center, Peking University \\
\textsuperscript{3}National Biomedical Imaging Center, College of Future Technology, Peking University
}

\iclrfinalcopy % Uncomment for camera-ready version, but NOT for submission.
\begin{document}

\maketitle

\begin{abstract}
Concept Bottleneck Models (CBMs) provide inherent interpretability by first predicting a set of human-understandable concepts and then mapping them to labels through a simple classifier. While users can intervene in the concept space to improve predictions, traditional CBMs typically employ a fixed linear classifier over concept scores, which restricts interventions to manual value adjustments and prevents the incorporation of new concepts or domain knowledge at test time. These limitations are particularly severe in unsupervised CBMs, where concept activations are often noisy and densely activated, making user interventions ineffective. We introduce Chat-CBM, which replaces score-based classifiers with a language-based classifier that reasons directly over concept semantics. By grounding prediction in the semantic space of concepts, Chat-CBM preserves the interpretability of CBMs while enabling richer and more intuitive interventions, such as concept correction, addition or removal of concepts, incorporation of external knowledge, and high-level reasoning guidance. Leveraging the language understanding and few-shot capabilities of frozen large language models, Chat-CBM extends the intervention interface of CBMs beyond numerical editing and remains effective even in unsupervised settings. Experiments on nine datasets demonstrate that Chat-CBM achieves higher predictive performance and substantially improves user interactivity while maintaining the concept-based interpretability of CBMs.
\end{abstract}

\section{Introduction}

\begin{wrapfigure}{tr}{0.5\linewidth} % r表示靠右，宽度设为0.5行宽
  \centering
  \includegraphics[width=0.95\linewidth]{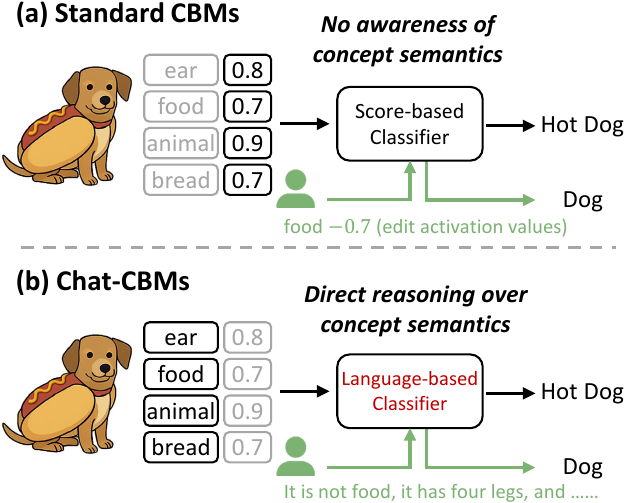}
  \caption{Illustration of standard CBMs and Chat-CBMs with score-/language-based classifiers.}
  \label{fig:method_overview}
\end{wrapfigure}

With the widespread adoption of deep learning, there is a growing demand for models that are both interpretable and interactive. This need is particularly critical in domains requiring trustworthy models, such as medical applications~\citep{explainable_artificial_review}, and in human-centered workflows requiring interactive and controllable models~\citep{Ilastik_nature_method, explanation_in_interactive}. Post-hoc explanation~\citep{DARPA_xai_program} methods attempt to rationalize model predictions through techniques such as feature attribution~\citep{attribution_review} and concept-based explanations~\citep{concept_based_exp}. However, their reliability is often questioned: potential biases in the explanation process make it difficult to separate flaws in the underlying model from artifacts of the explanation method itself~\citep{rudin2019stop}. Concept bottleneck models (CBMs)~\citep{cbm_icml2020}, in contrast, are interpretable models by design, which first map inputs to a set of human-understandable concepts and then predict class labels through this concept bottleneck (Figure~\ref{fig:method_overview} (a)). Crucially, the concept bottleneck also acts as an intervention interface where users can adjust concept activations to steer predictions. This user intervention ability is the key essential of CBMs and distinguishes them from alternative interpretable architectures such as the CapsuleNet~\citep{capsule_hinton_2017} and ProtoPNet~\citep{ProtoPNet_neurips2018,ProtoPFormer}.

Like other interpretable models, CBMs are subject to the well-known trade-off between interpretability and accuracy~\citep{xai_tradeoff,cem_neurips2022}. Their predictive performance often falls short of black-box counterparts, limiting adoption in domains where accuracy cannot be compromised~\citep{accuracy_Not_fairness_NEJM}. To narrow this gap, recent work has explored richer concept representations, more sophisticated intervention mechanisms, and intervention-aware models~\citep{cem_neurips2022,ecbm_iclr2024,intervention_procedure_icml2023,scbm,leanring_to_intervene_icml2024,interactive_cbm_aaai2023}. Yet, most existing CBMs still rely on score-based label predictors, which restrict user interventions to numerical edits of concept scores and prevent the addition or removal of concepts at test time. These limitations are exacerbated in unsupervised CBMs~\citep{LFcBM,labo}, which typically leverage CLIP-based~\citep{clip} vision–language similarity over large concept banks. Lacking explicit supervision, such models often produce noisy, densely activated concept predictions~\citep{shortcut_learning,waffleCLIP}, undermining interpretability and rendering effective user intervention nearly impossible.

In this work, we argue that these challenges primarily stem from the reliance on score-based label predictors. We propose Chat-CBM, which shifts the inference paradigm from numerical concept activations to concept semantics by employing a language-based classifier as the CBM predictor. Chat-CBM integrates concept semantics directly into the prediction process: labels are inferred through reasoning over concept semantics rather than activation scores. This design preserves the core essentials of CBMs, the concept-based interpretability, while extending the range of possible interventions. As illustrated in Figure~\ref{fig:method_overview} (b), Chat-CBM enables intuitive, language-driven interventions that surpass simple score adjustments, including concept correction, addition or removal of concepts, and high-level reasoning guidance. We conduct extensive experiments across nine datasets to evaluate Chat-CBM. Our results show that it outperforms traditional CBMs in classification accuracy, offers conversational interventions, and exhibits effective interventions even for unsupervised CBMs. These findings highlight the promise of language-based label predictors for building more interactive CBMs.

\section{Related Work}
\subsection{Concept Bottleneck Models}
\paragraph{Supervised CBMs.} 
For datasets with annotated concept labels, research on CBMs has primarily focused on enhancing concept representations and strengthening their intervention capabilities. For instance, CEM~\citep{cem_neurips2022} replaces scalar concept logits with learnable positive/negative embeddings; ProbCBM~\citep{probcbm_icml2023} introduces probabilistic concept embeddings to capture uncertainty; ECBM~\citep{ecbm_iclr2024} employs energy-based functions over (input, concept, label) triplets to better model joint dependencies; and SCBM~\citep{scbm} uses multivariate Gaussian distributions to represent correlated concept predictions. In parallel, new training paradigms and intervention policies have been explored. Interactive CBM~\citep{interactive_cbm_aaai2023} introduces the CooP policy, which estimates concept uncertainty to decide when user input should be requested, while IntCEM~\citep{learning_to_reveive_help_neurips2023} trains the model to actively select which concepts to query at inference. Recent works also investigate interventions, deployment under distribution shift~\citep{intervention_under_ood,he2025training}, and robustness to label noise~\citep{cpo_concept_mislabeling,editable_cbms}. Despite these advancements, supervised CBMs still rely on score-based classifiers for label prediction, which fundamentally restricts the flexibility of user interventions.

\paragraph{Unsupervised CBMs.} 
Obtaining fine-grained concept annotations is costly and often infeasible. To mitigate this, unsupervised CBMs have emerged, typically by constructing a concept bank using LLMs~\citep{llm_fewshot_learner} or vision–language models~\citep{bhalla2024interpreting}, followed by different concept filtering strategies~\citep{LFcBM,labo,LM4CV,v2c_cbm,coarse2fine_cbm,opencbms,xie2025discovering}. A label predictor is subsequently trained on concept activations computed by image–text similarity. While these approaches successfully reduce annotation cost, they suffer from limited intervention capability and are difficult to integrate with advanced CBM architectures. Moreover, the large number of concepts in the bank and the dense, noisy activations produced by CLIP features hinder the identification of actionable concepts~\citep{waffleCLIP}, making interventions ineffective and difficult to scale.

\subsection{Concept-based Interpretable Reasoning beyond Linear Classifiers}
Beyond CBMs, several other interpretable architectures also reason over concepts. For instance, DCR constructs syntactic rule structures using concept embeddings~\citep{barbiero2023interpretable}, while CMR supports more logic-driven decision processes and enables rule-based interventions~\citep{debot2024interpretable}. Prototype-based networks learn concept prototypes~\citep{ProtoPNet_neurips2018}, but the semantics of these prototypes require post-hoc analysis and, critically, do not support user interventions. XBM leverages multimodal LLMs to generate captions as concepts and then trains a BERT~\citep{bert} for downstream prediction, reducing annotation requirements but still suffering from low intervention efficiency~\citep{xbm}. Recent works have attempted to integrate concept bottleneck structures into LLMs. CB-LLM transforms a standard LLM into a CBM framework for interpretable text classification~\citep{sun2024concept}, while CB-pLM follows a similar strategy but is used for protein design~\citep{ismail2024concept}. In contrast, our proposed Chat-CBM emphasizes the semantic information inherently available in the concept bottleneck, and importantly, explores language-based interventions that are flexible and effective for both supervised and unsupervised CBMs.

\begin{figure}[t]
    \centering
    \includegraphics[width=1.0\linewidth]{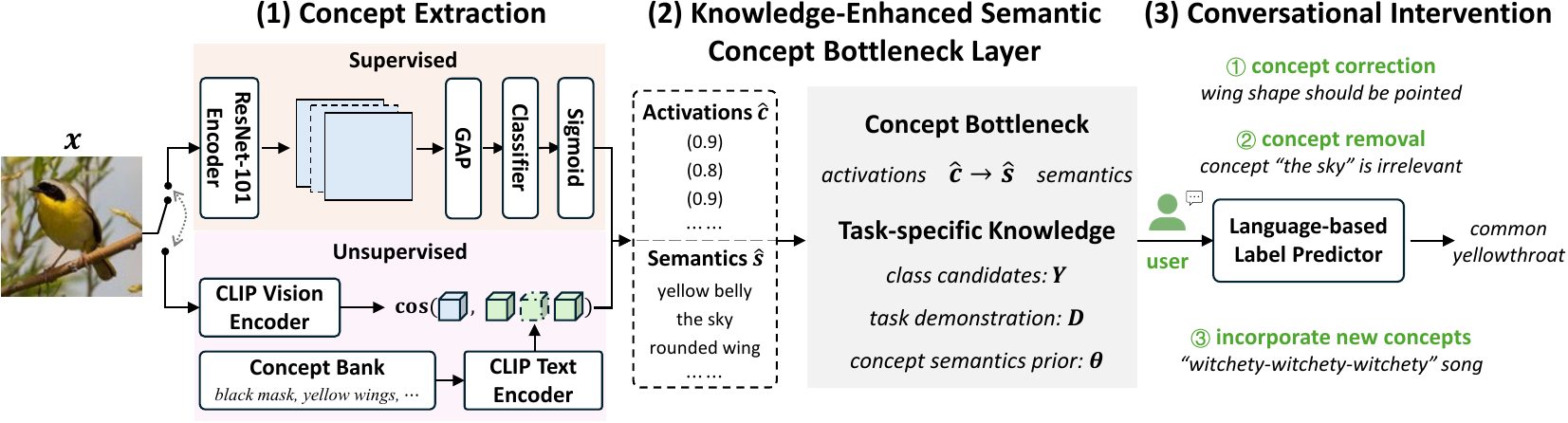}
    \caption{Overview of Chat-CBMs. We first extract concept semantics from the input images, then generate class candidates $\boldsymbol{Y}$ using the baseline CBMs and incorporate the task demonstration $\boldsymbol{D}$ and the concept semantics prior $\boldsymbol{\theta}$ of the class candidates to form the knowledge-enhanced semantic concept bottleneck layer. Finally, the language-based label predictor $f_{\cal M}$ reasons directly in the semantic bottleneck space, producing the final predictions while supporting flexible user interventions.}
    \label{fig:method_details}
\end{figure}

\section{Method}
\subsection{Problem Definition and Method Overview}
To overcome the limitations of score-based classifiers, where concept semantics are neglected and interventions are restricted to manual edits of activation values, we propose \textbf{Chat-CBM}, an interactive CBM that employs a language-based classifier $f_{\cal M}$ for label prediction over a \emph{knowledge-enhanced semantic concept bottleneck layer}. Unlike CBMs that operate in a numeric bottleneck of concept scores, Chat-CBM performs prediction in the semantic space of the concept bottleneck, maintaining core concept-based interpretability while extending its intervention ability.

As shown in Figure~\ref{fig:method_details}, Chat-CBM first obtains concept predictions $(\hat{\boldsymbol{c}}, \hat{\boldsymbol{s}})$ from the input $\boldsymbol{x}$, where $\hat{\boldsymbol{c}} \in [0,1]^{N_c}$ denotes the activation scores of $N_c$ concepts, and $\hat{\boldsymbol{s}}$ encodes their corresponding concept semantics. These are integrated into the bottleneck layer, together with task-specific knowledge. This includes class candidates $\boldsymbol{Y}=\{\boldsymbol{y}_i\}$ computed with a linear classifier $f(\hat{\boldsymbol{c}})$, task demonstration $\boldsymbol{D}$, and concept semantics prior $\boldsymbol{\theta}$ for each class. A frozen LLM $f_{\cal M}$ then computes the label prediction by selecting the candidate $\boldsymbol{y}_i$ with the highest probability conditioned on $\boldsymbol{D}$ and $\hat{\boldsymbol{s}}$:
\begin{equation}
    P(\boldsymbol{y}_i \mid \boldsymbol{D}, \boldsymbol{\theta}, \hat{\boldsymbol{s}}) \triangleq f_{\cal M}(\boldsymbol{y}_i, \boldsymbol{D}, \boldsymbol{\theta}, \hat{\boldsymbol{s}}), \qquad 
    \hat{\boldsymbol{y}} = \mathop{\arg\max}_{\boldsymbol{y}_i \in \boldsymbol{Y}} \ P(\boldsymbol{y}_i \mid \boldsymbol{D}, \boldsymbol{\theta}, \hat{\boldsymbol{s}}).
\end{equation}

By explicitly situating prediction within the knowledge-enhanced semantic concept bottleneck layer, Chat-CBM preserves the interpretability-by-design property of CBMs while enabling richer and more flexible user interventions, including standard concept correction as well as flexible concept removal or integration at test time. Details of each stage are described as follows.

% chat-cbm section
\subsection{Chat-CBM}\label{section:method_chat_cbm}
\subsubsection{Concept Extraction}
As shown in Figure~\ref{fig:method_details} (1), our method supports both datasets with and without concept annotations by adapting to either supervised or unsupervised CBMs for concept extraction.
\paragraph{Supervised CBMs.}
Given a dataset with concept labels, denoted as ${\cal D}=\{(\boldsymbol{x}^{(i)}, \boldsymbol{c}^{(i)}, \boldsymbol{y}^{(i)})\}$, where the $i$-th data point contains input $\boldsymbol{x}^{(i)}\in{\cal X}$, concept label $\boldsymbol{c}^{(i)}\in {\cal C}=\{0,1\}^{N_c}$, and one-hot class label $\boldsymbol{y}^{(i)}\in {\cal Y}=\{0,1\}^M$ of $M$ classes. As shown in Figure~\ref{fig:method_details} (orange), a supervised CBM is composed of a concept predictor $g: {\cal X}\rightarrow {\cal C}$ and a class predictor $f: {\cal C} \rightarrow {\cal Y}$. To mitigate the possible concept leakage problem and improve intervention efficiency~\citep{concept_leakage}, we train the concept predictor $g(\cdot)$ and label predictor $f(\cdot)$ independently. The concept activation values $\hat{\boldsymbol{c}}$ and semantics $\hat{\boldsymbol{s}}$ are obtained by:
\begin{gather}
    \hat{\boldsymbol{c}} = g(\boldsymbol{x}),\ \ \ \ \ \ \ \ \ \hat{\boldsymbol{s}} = {\rm decode}(\hat{\boldsymbol{c}}),
\end{gather}
where ${\rm decode}(\cdot)$ returns the concept semantics when activation values are larger than 0.5.
\paragraph{Unsupervised CBMs.}
For datasets ${\cal D}=\{(\boldsymbol{x}^{(i)}, \boldsymbol{y}^{(i)})\}$ without concept annotations, we build unsupervised CBMs for concept extraction. As shown in Figure~\ref{fig:method_details}, we first adopt the concept bank ${\cal B}=\{t_1,\cdots,t_{N_c}\}$ from~\citep{labo,v2c_cbm}, which consists of $N_c$ concepts. Then, we leverage CLIP to encode visual features of the input image and textual features of each concept in the concept bank, and compute the cosine similarity as the concept activation values $\hat{\boldsymbol{c}}$:
\begin{gather}
    \hat{\boldsymbol{c}} = g(\boldsymbol{x}) = \cos (e_v(\boldsymbol{x}), e_t({\cal B})) \in {\mathbb R}^{N_c}, \label{eq:unsupervised_cbm_c_pred}
\end{gather}
where $e_v$ and $e_t$ are CLIP vision and text encoders, respectively. Then the semantics of the top-10 activated concepts are used by the label predictor. (The top-N choice is discussed in appendix~\ref{appendix:training_details})

\subsubsection{Knowledge-Enhanced Semantic Concept Bottleneck Layer}
As illustrated in Figure~\ref{fig:method_details} (2), we then construct the knowledge-enhanced semantic concept bottleneck, including the concept semantics $\hat{\boldsymbol{s}}$, the class candidates \textbf{\textit{Y}}, a demonstration set \textbf{\textit{D}} consists of in-context learning (ICL) examples, and integrate the concept semantics prior $\boldsymbol{\theta}$ for the candidate classes.
\paragraph{Class Candidates Generation.}
In order to obtain the class candidates $\boldsymbol{Y}$ for Chat-CBM, we use the label predictor $f(\cdot)$ of the baseline CBM and take the top-N predictions as $\boldsymbol{Y}$:
\begin{equation}
    \boldsymbol{Y} = \{\boldsymbol{y}_1,\cdots,\boldsymbol{y}_N\} = \operatorname{top\text{-}N}(f(\hat{\boldsymbol{c}})) \label{eq:answer_candidates}.
\end{equation}
\paragraph{In-Context Learning Examples Selection.}
To enhance the reasoning capability of Chat-CBMs with frozen LLMs, we employ ICL to encourage LLMs to learn the associations between concepts and labels in the demonstration and, accordingly, make the right prediction. For each answer candidate $\boldsymbol{y}_i$ in $Y$, we randomly select $K$ samples of class $\boldsymbol{y}_i$ from the val set with their predicted concept semantics $\hat{\boldsymbol{s}}_{\rm val}$ to form the ICL demonstrations:
\begin{gather}
    \boldsymbol{D} = \{ I, (\hat{\boldsymbol{s}}^{(1)}_{{\rm val}}, \boldsymbol{y}_1),\cdots,(\hat{\boldsymbol{s}}^{(K)}_{{\rm val}}, \boldsymbol{y}_1),\cdots,(\hat{\boldsymbol{s}}^{(K)}_{{\rm val}}, \boldsymbol{y}_N)\},
\end{gather}
where $I$ is the task instruction with format control like ``\textit{Answer the image class based on the concepts, the answer format is $<$analysis: ...,$>$ $<$answer: ...$>$}".
%and we use the demonstration set $\boldsymbol{D}$ as the $N$-way $K$-shot settings for image classification with Chat-CBMs.

\paragraph{Class Concept Semantics Prior Integration.}
Beyond enhancing the local mapping relationships between concepts and class labels via ICL, another advantage of using LLMs is that they also allow the incorporation of global task-specific knowledge. To further enrich reasoning, we optionally augment the input with structured class prior knowledge $\boldsymbol{\theta}$, which describes the most common attributes for each candidate class $\boldsymbol{y}_i$. The inference objective then becomes:
\begin{equation}
    \hat{\boldsymbol{y}} = \mathop{\arg\max}_{\boldsymbol{y}_i \in \boldsymbol{Y}} \ P(\boldsymbol{y}_i \mid \boldsymbol{D}, \boldsymbol{\theta},\hat{\boldsymbol{s}}),
\end{equation}
where $\theta$ serves as an additional global class prior, and can take various forms, including natural language descriptions, hierarchical taxonomies, visual-textual traits, and so on. 

\begin{figure}[tb]
    \centering
    \includegraphics[width=0.9\linewidth]{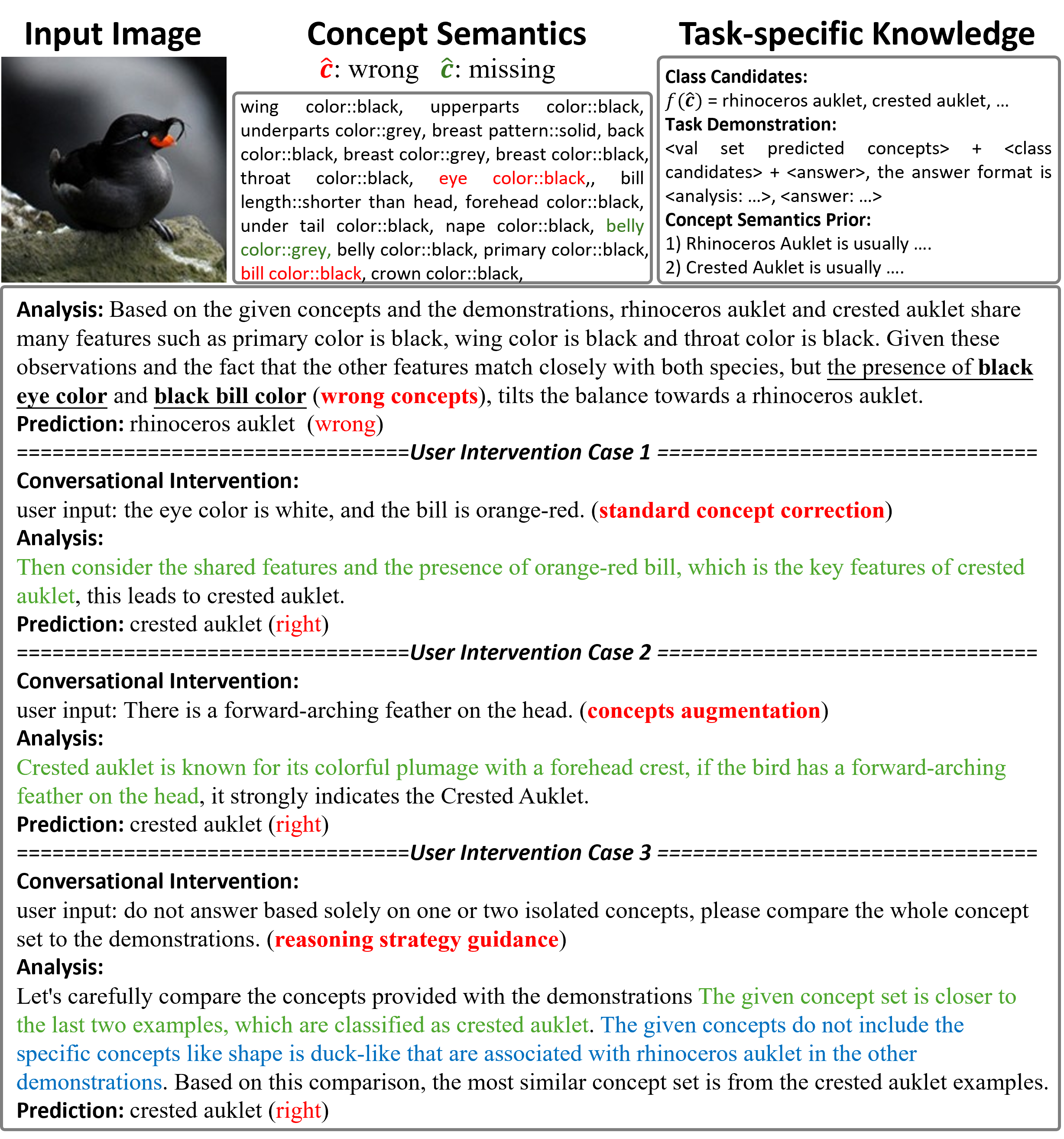}
    \caption{Conversational intervention on the CUB dataset. \textcolor{green_color}{Green} highlights the positive reasoning and \textcolor{blue_color}{blue} highlights the negative reasoning process. Users can either directly correct concept predictions like standard CBMs (\textbf{case 1}), adding new (or removing) concepts beyond the predefined concept bottleneck (\textbf{case 2}), or give a high-level reasoning strategy to guide thinking (\textbf{case 3}).}
    \label{fig:visualization_cub_intervention_1}
\end{figure}

\subsubsection{Intervention}
\paragraph{Standard Numerical Intervention.}
Chat-CBMs retain the standard intervention abilities of CBMs, allowing users to directly edit concept activation values ($u(\cdot)$ denotes user intervention). Given updated activations $\hat{\boldsymbol{c}}_{\rm new}=u(\hat{\boldsymbol{c}})$, the corresponding concept semantics $\hat{\boldsymbol{s}}_{\rm new}$ and class candidates $\boldsymbol{Y}_{\rm new}$ are passed to the language-based classifier for inference.

\paragraph{Conversational Intervention.} 
Beyond numerical edits, the language-based classifier enables flexible interventions via natural language $u_{\rm text}$. We highlight three representative types of conversational interventions and provide examples in Figure~\ref{fig:visualization_cub_intervention_1}:  
\begin{itemize}[noitemsep, topsep=0pt, leftmargin=*]
  \item \textbf{Concept correction}: Standard corrections can also be performed conversationally with awareness of prior reasoning, e.g., ``\textit{the concept forest is wrongly predicted}''.
  \item \textbf{Concept augmentation/removal}: New concepts $ {\boldsymbol{s}}_{\text{new}} \notin \hat{\boldsymbol{s}} $ can be added, or existing ones ${\boldsymbol{s}}_i \in \hat{\boldsymbol{s}}$ removed, via prompts such as ``\textit{the bird also has a forward-arching feather on the head}'' or ``\textit{ignore concepts about bird size during analysis}''.
  \item \textbf{High-level strategy guidance}: Users can provide high-level reasoning strategies, e.g., ``\textit{focus on the bird size when distinguishing common yellowthroat and yellow-breasted chat}''.
\end{itemize}
Formally, intervention messages $u_{\rm text}$ are incorporated into the conversation history $\cal H$, and new predictions are generated as
\begin{equation}
\hat{\boldsymbol{y}}_{\rm new} = \arg\max_{\boldsymbol{y}_i \in Y} P(\boldsymbol{y}_i \mid \boldsymbol{D}, \boldsymbol{\theta}, \hat{\boldsymbol{s}}, {\cal H}, u_{\rm text}).
\end{equation}
As shown in Figure~\ref{fig:visualization_cub_intervention_1}, Chat-CBM naturally combines positive reasoning (e.g., ``\textit{forehead-arching feather is distinctive for Crested Auklet}'') and negative reasoning (e.g., ``\textit{the given concepts lack duck-like shape features for Rhinoceros Auklet}''). This process allows users to understand and take control of the decision pipeline.

\begin{table}[!t]
\centering
\caption{Classification accuracy on datasets with concept labels. We report the mean and standard deviation from five runs with different random seeds. (LLaMA-3-70B-Instruct for Chat-CBM.)}
\label{classification_accuracy}
\resizebox{\linewidth}{!}{
\begin{tabular}{l|cc|cc|cc}

\toprule
\diagbox{\textbf{Model}}{\textbf{Data}} & \multicolumn{2}{c|}{\textbf{CUB}}  & \multicolumn{2}{c|}{\textbf{AwA2}}  & \multicolumn{2}{c}{\textbf{PBC}}\\ 
\midrule
Metric & Concept Acc. & Class Acc. & Concept Acc. &  Class Acc. & Concept Acc. & Class Acc.\\ 
\midrule
End-to-End & - & 0.825 $\pm$ 0.002 & - & 0.953 $\pm$ 0.001 & - & 0.997 $\pm$ 0.000\\
\hline
Hard CBM & 0.960 $\pm$ 0.004 & 0.708 $\pm$ 0.003 & 0.980 $\pm$ 0.001 & 0.901 $\pm$ 0.001 & 0.920 $\pm$ 0.005 & 0.959 $\pm$ 0.011 \\
ProbCBM & 0.955 $\pm$ 0.003 & 0.723 $\pm$ 0.001 & 0.959 $\pm$ 0.000 & 0.890 $\pm$ 0.007 & 0.950 $\pm$ 0.001 & 0.990 $\pm$ 0.002 \\ 
CEM  & 0.962 $\pm$ 0.002 & 0.799 $\pm$ 0.003 & 0.979 $\pm$ 0.003 & 0.924 $\pm$ 0.002 & 0.952 $\pm$ 0.002 & 0.993 $\pm$ 0.001\\ 
CBM & 0.965 $\pm$ 0.009 & 0.752 $\pm$ 0.005 & 0.982 $\pm$ 0.000 & 0.923 $\pm$ 0.004 & 0.956 $\pm$ 0.003 & 0.988 $\pm$ 0.008 \\
%\rowcolor{gray!20}
+ \textbf{Chat-CBM} & 0.965 $\pm$ 0.009 & 0.815 $\pm$ 0.005 & 0.982 $\pm$ 0.000 & \textbf{0.964} $\pm$ 0.002 & 0.956 $\pm$ 0.003 & 0.986 $\pm$ 0.002 \\
ECBM & 0.967 $\pm$ 0.003 & 0.806 $\pm$ 0.004 & 0.983 $\pm$ 0.001 & 0.916 $\pm$ 0.000 & 0.935 $\pm$ 0.004 & \textbf{0.994} $\pm$ 0.001 \\ 
%\rowcolor{gray!20}
+ \textbf{Chat-CBM} & 0.967 $\pm$ 0.003 & \textbf{0.816} $\pm$ 0.006 & 0.983 $\pm$ 0.001 & 0.961 $\pm$ 0.005 & 0.935 $\pm$ 0.004 & 0.989 $\pm$ 0.001\\
\bottomrule
\end{tabular}
}
\vspace{-0.3cm}
\end{table}

\section{Experiments}\label{section:experiment}
%\subsection{Setup}
\paragraph{Datasets.} 
We employed two types of datasets to validate the effectiveness of our approach in both supervised and unsupervised CBMs. Datasets with concept labels: (1) CUB~\citep{cub_dataset}, a fine-grained bird classification dataset, we follow~\citep{cbm_icml2020} to use 112 concepts, (2) AwA2~\citep{awa2_dataset}, which contains 50 animal classes with 85 attributes, and (3) PBC~\citep{PBCdataset}, a white blood cell classification dataset with 5 white blood cell classes and 11 morphological attributes (31 concepts) from~\citep{wbcatt_dataset}. Datasets without concept labels: (1) DTD~\citep{texture_dataset} for abstract texture classification of 47 classes, (2) Food-101~\citep{food101_dataset} with 101 types of food, (3) Flower-102~\citep{flower_102} for fine-grained classification of 102 types of flowers, and (4) CIFAR10, (5) CIFAR100~\citep{cifar_dataset}, (6) ImageNet~\citep{imagenet_dataset} as standard classification benchmarks. For all datasets, we use the same data split settings for training and evaluating the performance of different methods.
\paragraph{Implementation Details.} We compare our Chat-CBM to both \textbf{(1)} supervised CBMs, including CBM~\citep{cbm_icml2020}, Hard CBM which uses 0/1 activation values for CBM~\citep{concept_leakage}, ProbCBM~\citep{probcbm_icml2023}, CEM~\citep{cem_neurips2022}, and ECBM~\citep{ecbm_iclr2024} and \textbf{(2)} unsupervised CBMs such as LaBo~\citep{labo} and V2C-CBM~\citep{v2c_cbm}. We use ResNet-101~\citep{resnet} as the backbone for supervised CBMs and CLIP ViT-L/14~\citep{clip} for unsupervised CBMs. And we test Chat-CBMs with LLaMA3-Instruct~\citep{llama3} and Qwen2.5-Instruct~\citep{Qwen} as the language-based classifiers. We use AdamW~\citep{adamw} and ConsineAnnealingLR for training all baseline models. All images are resized to 224$\times$224 for both training and testing. We use the same training settings for each dataset and report the mean and standard deviation across five runs with different random seeds. Full details on training and evaluation are provided in appendix~\ref{appendix:all_details}.

\subsection{Classification Performance}
\begin{figure}[!t]
    \centering
    \includegraphics[width=0.95\linewidth]{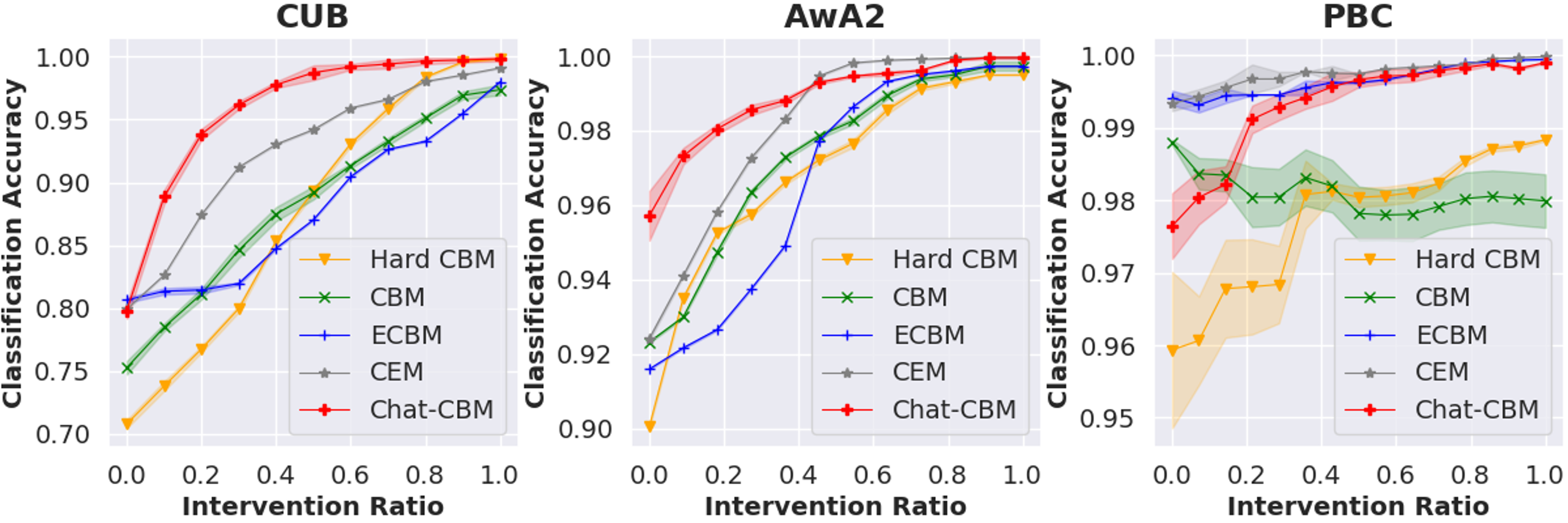}
    \caption{Intervention via Concept Correction. The LLM for Chat-CBM is LLaMA3-8B-Instruct.\vspace{-0.3cm}}
    \label{fig:intervention_on_dataset_with_concept_labels}
    \vspace{-0.2cm}
\end{figure}

\paragraph{Compared to Supervised CBMs.} Table~\ref{classification_accuracy} presents the classification accuracy of different methods on datasets with concept labels. Remarkably, even with frozen LLMs, our Chat-CBM surpasses the baselines on the CUB and AwA2 datasets (the top-N classification performance of baselines is in Table~\ref{table:topk_classification_accuracy}). For the PBC dataset, we found that standard CBMs~\citep{cbm_icml2020}, although exhibit high classification accuracy, suffer from severe concept leakage problems~\citep{concept_leakage}, because the intervention procedure is ineffective for them as shown in Figure~\ref{fig:intervention_on_dataset_with_concept_labels}, but our Chat-CBM can achieve better performance compared with Hard CBMs~\citep{concept_leakage} and also show an effective intervention curve. We also test our Chat-CBMs with ECBMs~\citep{ecbm_iclr2024} as baselines, and Chat-CBMs achieve better classification performance due to the improvement of the baseline. 
\paragraph{Compared to Unsupervised CBMs.} For datasets without concept labels, the results are shown in Table~\ref{classification_accuracy_without_concept_label}. While the performance of Chat-CBM is inferior to LaBo and V2C-CBM under the all-shot setting, this gap is largely due to the noisy concept bottlenecks in VL-CBMs, while Chat-CBM leverages frozen LLMs without any fine-tuning. So we mainly compare Chat-CBMs against the 2-shot performance of the baselines. In summary, Chat-CBM with Qwen2.5-32B-Instruct achieves an average improvement of 1.83\% over LaBo and 9.53\% over V2C-CBM across the six datasets. Interestingly, Chat-CBMs with V2C-CBMs as baselines demonstrate better performance than LaBo baselines, and we think this is because the concept bank of V2C-CBMs contains more accurate and concise visual concepts compared to LaBo, as discussed in~\citep{v2c_cbm}.
\begin{table}[!t]
\centering
\caption{Classification accuracy on datasets without concept labels. We report the mean and standard deviation from five runs with different random seeds. (Qwen2.5-32B-Instruct for Chat-CBM)}
\label{classification_accuracy_without_concept_label}
\resizebox{\linewidth}{!}{
\begin{tabular}{l|c|c|c|c|c|c}

\toprule
\diagbox[width=12em]{\textbf{Model}}{\textbf{Data}} &    \textbf{DTD}
&\textbf{Food-101} 
&\textbf{Flower-102} 
&\textbf{CIFAR10} & \textbf{CIFAR100} & \textbf{ImageNet} \\ 
\midrule
Linear Prob (All)    &    0.821 $\pm$ 0.003 
&0.952 $\pm$ 0.000 
&0.993 $\pm$ 0.001 
&0.981 $\pm$ 0.000 & 0.873 $\pm$ 0.001 & 0.841 $\pm$ 0.003 \\
Linear Prob (1-shot) &    0.436 $\pm$ 0.010 
&0.578 $\pm$ 0.004 
&0.477 $\pm$ 0.003 
&0.624 $\pm$ 0.003 & 0.393 $\pm$ 0.008 & 0.422 $\pm$ 0.004 \\
Linear Prob (2-shot) &    0.537 $\pm$ 0.001 
&0.749 $\pm$ 0.001 
&0.610 $\pm$ 0.003 
&0.803 $\pm$ 0.002 & 0.574 $\pm$ 0.003 & 0.558 $\pm$ 0.005 \\
\hline
LaBo (All)    &    0.769 $\pm$ 0.001 
&0.924 $\pm$ 0.005 
&0.993 $\pm$ 0.001 
&0.978 $\pm$ 0.001 & 0.860 $\pm$ 0.002 & 0.840 $\pm$ 0.006 \\
LaBo (1-shot) &    0.531 $\pm$ 0.016 
&0.806 $\pm$ 0.009 
&0.825 $\pm$ 0.003 
&0.910 $\pm$ 0.002 & 0.627 $\pm$ 0.007 & 0.512 $\pm$ 0.014 \\
LaBo (2-shot) &    0.552 $\pm$ 0.004 
&0.840 $\pm$ 0.002 
&0.895 $\pm$ 0.001 
&0.910 $\pm$ 0.001 & 0.658 $\pm$ 0.003 & 0.571 $\pm$ 0.008 \\
%\rowcolor{gray!20}
\textbf{LaBo-Chat-CBM} (2-shot) &    0.677 $\pm$ 0.011 
&0.753 $\pm$ 0.003 
&0.876 $\pm$ 0.002 
&0.889 $\pm$ 0.002 & 0.670 $\pm$ 0.004 & 0.601 $\pm$ 0.002 \\
\hline 
V2C-CBM (All)    &    0.782 $\pm$ 0.003 
&0.927 $\pm$ 0.002 
&0.987 $\pm$ 0.002 
&0.980 $\pm$ 0.000 & 0.864 $\pm$ 0.000 & 0.841 $\pm$ 0.002 \\
V2C-CBM (1-shot) &    0.421 $\pm$ 0.017 
&0.586 $\pm$ 0.024 
&0.884 $\pm$ 0.009 
&0.893 $\pm$ 0.008 & 0.627 $\pm$ 0.015 & 0.561 $\pm$ 0.009 \\
V2C-CBM (2-shot) &    0.492 $\pm$ 0.003 
&0.745 $\pm$ 0.005 
&0.930 $\pm$ 0.009 
&0.934 $\pm$ 0.002 & 0.651 $\pm$ 0.003 & 0.615 $\pm$ 0.005 \\
% \rowcolor{gray!20}
\textbf{V2C-Chat-CBM} (2-shot) & 0.734 $\pm$ 0.004 & 0.786 $\pm$ 0.019 & 0.914 $\pm$ 0.002 & 0.955 $\pm$ 0.007 & 0.727 $\pm$ 0.002 & 0.667 $\pm$ 0.004 \\
\bottomrule
\end{tabular}
}
\vspace{-0.1cm}
\end{table}

\subsection{Intervention}
\begin{figure}[!t]
    \centering
    \includegraphics[width=1.0\linewidth]{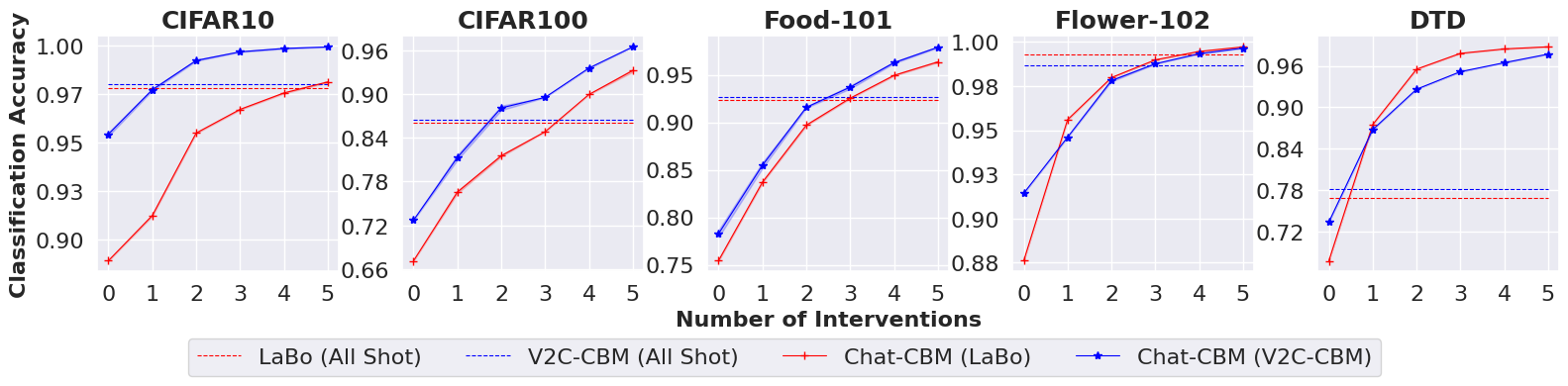}
    \caption{Intervention on datasets without concept labels. The LLM for Chat-CBM is Qwen2.5-32B-Instruct with the same setting in Table~\ref{classification_accuracy_without_concept_label}. The all-shot baseline performance is provided for reference because of the infeasibility of conducting automated interventions for existing unsupervised CBMs. % The x-axis denotes the number of concepts intervened, and the y-axis denotes the classification accuracy.
    }
    \label{fig:intervention_unsupervised_datasets}
    
\end{figure}

\paragraph{Concept Correction via Standard Numerical Intervention.} We begin by evaluating standard concept correction on datasets with annotated concepts, following~\citep{cbm_icml2020,ecbm_iclr2024}. Specifically, we intervene on the baseline CBMs and update $\hat{\boldsymbol{s}}$ and $\boldsymbol{Y}$ accordingly for Chat-CBM. The results in Figure~\ref{fig:intervention_on_dataset_with_concept_labels} highlight the effectiveness of Chat-CBM interventions, particularly on the CUB dataset, where fine-grained distinctions rely heavily on specific concepts. On the PBC dataset, independent CBMs suffer from severe concept leakage, which undermines intervention effectiveness. In contrast, Chat-CBM reasons directly in the semantic space of concepts through its language-based classifier, rather than relying on raw activation values, which prevents the label predictor from exploiting spurious class proxies and thereby yields consistent performance gains under intervention.
\paragraph{Intervention on Datasets without Concept Labels.} Direct user interventions on existing unsupervised CBMs are nearly infeasible. These models typically rely on hundreds or even thousands of concepts for reasoning, making it impractical to identify actionable concepts. A few prior studies have attempted to replace visual concepts with image captions for intervention, but only achieved marginal improvements~\citep{xbm}, highlighting the lack of effective intervention capabilities in current unsupervised CBMs. To automatically conduct interventions for Chat-CBMs, we employ an assistant LLM that selects concepts to emphasize, remove, or augment based on the conversation history of Chat-CBM, the top-20 predicted concepts, and the ground-truth class label. We use the top-10 label predictions as class candidates, ensuring a high accuracy upper bound. As shown in Figure~\ref{fig:intervention_unsupervised_datasets}, the x-axis denotes the number of interventions, with the assistant LLM restricted to editing only one concept from the top-20 at each step. Remarkably, Chat-CBM surpasses the all-shot performance of baselines within five interventions, demonstrating the feasibility and the effectiveness of language-based reasoning for enabling user interventions in unsupervised CBMs.

\begin{wrapfigure}{r}{0.5\linewidth} % r表示靠右，宽度设为0.5行宽
  \centering
  \includegraphics[width=1.0\linewidth]{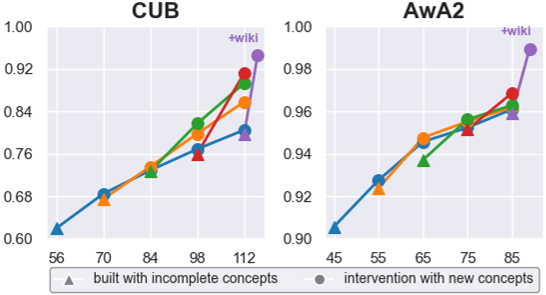}
  \caption{Intervention with new concepts (Chat-CBM with LLaMA3-8B-Instruct).}
  \label{fig:intervention_using_new_concepts}
\end{wrapfigure}

\paragraph{Intervention Using New Concepts.} We further test the situation when new concepts are introduced during test time. We first design a controlled setting where we train CBMs on the CUB and AwA2 datasets with incomplete concepts, and the rest are used as new concepts for intervention. The results are shown in Figure~\ref{fig:intervention_using_new_concepts}. Because some concepts are useful for classification, and may be easily recognized by users (such as the \textit{forward-arching feather on the head} in Figure~\ref{fig:visualization_cub_intervention_1} case 2), or can come from beyond the images (such as sounds and smell). We also explore the descriptions of the target class from Wikipedia and replace the class names with general names like ``the bird" or ``the animal", and then use them to intervene in Chat-CBMs. Although the huge performance improvement, as shown in Figure~\ref{fig:intervention_using_new_concepts} (+wiki), seems obvious because of the rich information from Wikipedia, we want to argue that previous activation-based classifiers do not support this type of intervention, and this remains a distinctive advantage of language-based CBMs (detailed implementations in appendix~\ref{appendix:wiki}). 

\subsection{Ablation Study}
\paragraph{Ablation on the Knowledge-Enhanced Semantic Concept Bottleneck.}
We further conduct an ablation study on diverse input configurations to validate the efficacy of our knowledge injection strategies for our semantic concept bottleneck, with quantitative results presented in Table~\ref{table:ablation_on_prior_knowledge_and_icl}. Strategy without $\boldsymbol{D}$-GT means that $\boldsymbol{D}$ also contains the class candidates $\boldsymbol{Y}_{\rm val}$ for the ICL examples. Strategy $\boldsymbol{\theta}$ means integrating dataset-specific class prior knowledge $\boldsymbol{\theta}$, with detailed implementations in appendix~\ref{prior_knowledge_appendix}. Our analysis reveals that incorporating prior knowledge yields significant improvements in Chat-CBM's classification accuracy. This enhancement stems from a key observation: while the concept predictor achieves reasonably high accuracy at the concept level, the overall concept accuracy remains suboptimal (as noted in~\cite{ecbm_iclr2024}). Individual concept prediction errors can consequently misguide Chat-CBM's final decisions. The introduced prior knowledge effectively enhances Chat-CBM's robustness against noise in concept predictions. We also examine how the number of ICL examples affects Chat-CBM's performance (Table~\ref{table:ablation_on_number_of_shots}). Chat-CBM demonstrates strong few-shot learning capabilities inherent to modern LLMs, with classification performance scaling positively with the number of ICL examples. However, we note that increasing ICL examples linearly expands context length, and further scaling ICL requires additional computational resources and larger LLM architectures to better leverage the information.

\begin{table}[t]
\centering
\caption{Ablation on knowledge enhancement strategies for the semantic concept bottleneck layer.}
\label{table:ablation_on_prior_knowledge_and_icl}
\resizebox{\linewidth}{!}{
\begin{tabular}{c|ccc|ccc|cc}
\toprule
\multirow{2}*{\textbf{Model}} & \multicolumn{3}{c|}{\textbf{Strategy}} & \multicolumn{3}{c}{\textbf{Supervised Dataset}} & \multicolumn{2}{c}{\textbf{Unsupervised Dataset}}\\
 & $\boldsymbol{D}$ & $\boldsymbol{D}$-GT & $\boldsymbol{\theta}$ & CUB & AwA2 & PBC & Flower-102 & DTD\\ 
\midrule
\multirow{4}*{\makecell{Chat-CBM\\(Qwen2.5-7B)}} 
 & \CheckmarkBold & & & 0.645 $\pm$ 0.007 & 0.722 $\pm$ 0.005 & 0.810 $\pm$ 0.007 & 0.853 $\pm$ 0.007 & 0.665 $\pm$ 0.022\\
 & \CheckmarkBold & \CheckmarkBold & & 0.655 $\pm$ 0.011 & 0.754 $\pm$ 0.006 & 0.830 $\pm$ 0.009 & 0.843 $\pm$ 0.021 & 0.629 $\pm$ 0.017\\
 & \CheckmarkBold & & \CheckmarkBold & 0.771 $\pm$ 0.008 & 0.817 $\pm$ 0.003 & 0.868 $\pm$ 0.003 & 0.862 $\pm$ 0.009 & 0.675 $\pm$ 0.009\\
 & \CheckmarkBold & \CheckmarkBold & \CheckmarkBold & 0.775 $\pm$ 0.013 & 0.871 $\pm$ 0.010 & 0.878 $\pm$ 0.007 & 0.899 $\pm$ 0.012 & 0.710 $\pm$ 0.011\\
\midrule
\multirow{4}*{\makecell{Chat-CBM\\(Qwen2.5-14B)}} & \CheckmarkBold & & & 0.734 $\pm$ 0.006 & 0.930 $\pm$ 0.019 & 0.899 $\pm$ 0.002 & 0.897 $\pm$ 0.005 & 0.689 $\pm$ 0.007\\
 & \CheckmarkBold & \CheckmarkBold & & 0.738 $\pm$ 0.011 & 0.932 $\pm$ 0.004 & 0.927 $\pm$ 0.004 & 0.848 $\pm$ 0.011 & 0.685 $\pm$ 0.015\\
 & \CheckmarkBold & & \CheckmarkBold & 0.776 $\pm$ 0.022 & 0.945 $\pm$ 0.004 & 0.949 $\pm$ 0.002 & 0.902 $\pm$ 0.003 & 0.701 $\pm$ 0.008 \\
 & \CheckmarkBold & \CheckmarkBold & \CheckmarkBold & 0.801 $\pm$ 0.003 & 0.951 $\pm$ 0.002 & 0.965 $\pm$ 0.003 & 0.910 $\pm$ 0.005 & 0.722 $\pm$ 0.013\\
\bottomrule
\end{tabular}
}
\vspace{-0.3cm}
\end{table}

\begin{table}[!t]
\centering
\caption{Ablation on the number of ICL examples.}
\label{table:ablation_on_number_of_shots}
\resizebox{\linewidth}{!}{
\begin{tabular}{l|ccc|ccc|ccc|ccc}
\toprule
\multirow{2}*{\diagbox[width=12em]{\textbf{Model}}{\textbf{Data}}} & \multicolumn{3}{c|}{\textbf{CUB}} & \multicolumn{3}{c|}{\textbf{AwA2}} & \multicolumn{3}{c|}{\textbf{WBC}} & \multicolumn{3}{c}{\textbf{Flower-102}}\\
 & N2-K1 & N2-K2 & N2-K3 & N2-K1 & N2-K2 & N2-K3 & N5-K1 & N5-K3 & N5-K5 & N2-K1 & N2-K2 & N2-K3 \\
\midrule
Chat-CBM (LLaMA3-8B)   & 0.784 & 0.797 & 0.798 & 0.910 & 0.957 & 0.965 & 0.930 & 0.976 & 0.980 & 0.893 & 0.915 & 0.914\\
Chat-CBM (Qwen2.5-7B)  & 0.735 & 0.775 & 0.782 & 0.871 & 0.871 & 0.898 & 0.827 & 0.878 & 0.913 & 0.856 & 0.899 & 0.912\\
Chat-CBM (Qwen2.5-14B) & 0.789 & 0.801 & 0.802 & 0.936 & 0.951 & 0.962 & 0.945 & 0.965 & 0.972 & 0.901 & 0.910 & 0.916\\
\bottomrule
\end{tabular}
}
\vspace{-0.3cm}
\end{table}

\begin{table}[!t]
\centering
\caption{Ablation on different LLMs and LLM sizes for Chat-CBMs.}\label{tab:ablation_on_LLMs}
\resizebox{\linewidth}{!}{
\begin{tabular}{lc|cc|cccc}
\toprule
\multicolumn{2}{c|}{\multirow{2}*{\diagbox[dir=SE,width=11em]{\textbf{Data}}{\textbf{Baseline}}{\textbf{LLM}}}} & \multicolumn{2}{c|}{\textbf{LLaMA3-Instruct}} & \multicolumn{4}{c}{\textbf{Qwen2.5-Instruct}}\\[0.5ex]
& & 8B & 70B & 7B & 14B & 32B & 72B\\
\midrule
CUB  & \multirow{3}*{\makecell{CBM}} & 0.797 $\pm$ 0.006 & \textbf{0.815} $\pm$ 0.005 & 0.775 $\pm$ 0.013 & 0.801 $\pm$ 0.003 & 0.803 $\pm$ 0.004 & \underline{0.812} $\pm$ 0.002 \\
AwA2 & & 0.957 $\pm$ 0.007 & \textbf{0.964} $\pm$ 0.002 & 0.871 $\pm$ 0.010 & \underline{0.951} $\pm$ 0.002 & 0.949 $\pm$ 0.002 & 0.950 $\pm$ 0.001 \\
PBC  & & 0.976 $\pm$ 0.010 & \textbf{0.986} $\pm$ 0.002 & 0.878 $\pm$ 0.007 & 0.965 $\pm$ 0.002 & 0.975 $\pm$ 0.001 & \underline{0.976} $\pm$ 0.001 \\
\midrule
CIFAR10 & \multirow{3}*{V2C-CBM} & 
0.929 $\pm$ 0.007 & \underline{0.951} $\pm$ 0.006 & 0.950 $\pm$ 0.005 & 0.951 $\pm$ 0.012 & 0.955 $\pm$ 0.007 & \textbf{0.956} $\pm$ 0.005 \\
Flower-102 &                     & 
0.915 $\pm$ 0.008 & \textbf{0.933} $\pm$ 0.002 & 0.899 $\pm$ 0.012 & 0.910 $\pm$ 0.005 & 0.914 $\pm$ 0.002 & \underline{0.921} $\pm$ 0.003 \\
DTD &                            & 
0.731 $\pm$ 0.013 & \textbf{0.757} $\pm$ 0.009 & 0.710 $\pm$ 0.011 & 0.722 $\pm$ 0.013 & \underline{0.734} $\pm$ 0.004 & 0.732 $\pm$ 0.009 \\
\bottomrule
\end{tabular}
}
\vspace{-0.3cm}
\end{table}

\paragraph{Ablation on Different LLMs and LLM Sizes.}
We further evaluate Chat-CBM with different LLM backbones and model sizes, as shown in Table~\ref{tab:ablation_on_LLMs}. Larger LLMs generally achieve better classification performance than their small counterparts. This can be attributed to their stronger ability to capture the in-context mappings and to leverage the provided class prior knowledge, which is crucial when the concept inputs are noisy. On the Qwen2.5-Instruct series, we observe that increasing the model size beyond 14B does not lead to further significant improvements. This suggests that the model capacity is no longer a limiting factor—i.e., 14B is already sufficient to encode the necessary task structure and priors. Combined with the prompt ablation results in Table~\ref{table:ablation_on_prior_knowledge_and_icl}, we hypothesize that ICL combined with class priors provides adequate supervision, and further gains rely more on the model's robustness to noisy or imperfect concept inputs than on increased parameter count.

\section{Conclusion and Limitations}\label{section:conclusion_and_limitations}
We introduce Chat-CBM, which replaces the score-based classifier of conventional CBMs with a language-based predictor operating in a semantic concept bottleneck. This design preserves the concept-based interpretability by explicitly keeping a concept bottleneck structure, while extending the intervention ability beyond numeric edits. Experiments on both annotated and unannotated datasets show that Chat-CBM improves classification accuracy, supports multiple forms of intervention, and scales with ICL examples and model size. Limitations remain: the use of semantic concept bottlenecks prevents the concept leakage problem, but also limits the representation ability under an incomplete concept situation. The use of LLMs introduces extra inference cost, and deployment in sensitive domains requires safeguarding against harmful knowledge encoded in LLMs. More detailed discussions are provided in Appendix~\ref{appendix:limitation}.
%\subsubsection*{Acknowledgments}
%Use unnumbered third level headings for the acknowledgments. All
%acknowledgments, including those to funding agencies, go at the end of the paper.

\bibliography{iclr2026_conference}
\bibliographystyle{iclr2026_conference}
\clearpage
\appendix
\section{All Implementation Details}\label{appendix:all_details}
\subsection{Training Details}\label{appendix:training_details}
For all datasets with concept labels, we use a ResNet-101 pretrained on ImageNet1k as the concept predictor backbone for all models, including CBM, ProbCBM, CEM, ECBM, and SCBM, and use a single layer as the label predictor except for those architectures with advanced designs (such as ProbCBM and ECBM). We use AdamW as the optimizer and CosineAnnealingLR as the learning rate scheduler for training all models. All images are resized to 224$\times$224 for both training and testing. For all baseline models on datasets with concept labels, we train the concept predictor for 150 epochs and the label predictor for 50 epochs. For training LaBo and V2C-CBM, we use the implementation of~\citep{labo} with the same hyperparameters as discussed in the appendix of V2C-CBM. The baseline models are trained using \texttt{PyTorch} and \texttt{transformers} library~\citep{wolf-etal-2020-transformers} with one NVIDIA RTX4090 Graphics card, and the inference of LLMs is conducted on NVIDIA L40 cards (1 card for LLaMA3-8B, Qwen2.5-7B, and Qwen2.5-14B, 2 cards for Qwen-2.5-32B, and 4 cards for Qwen2.5-72B and LLaMA3-70B).

\subsection{Evaluation Details of Chat-CBM}\label{appendix:evaluation_details}
To control the output format of LLMs, we employ in-context examples and task instructions. The expected response format is $<$analysis: $>$ $<$answer: class name$>$. Therefore, we determine whether the LLM provides a correct answer by directly matching <answer: target class name> within the LLM-generated response. We also checked the output of the model in advance and found that this kind of format requirement could be easily followed by LLMs. When prompting LLMs with the \texttt{transformers} library, we set the following hyperparameters for all LLMs: \texttt{max\_length=8192}, \texttt{do\_sample=true}, and \texttt{top\_k=10}. We use left padding for the tokenizers and also set \texttt{max\_length=8192}. But for intervention experiments, we set the \texttt{max\_length=10240} because the input length may exceed 8192 after several turns of intervention.

\subsection{Top-N Classification Accuracy of CBMs}
Since we use a standard independent CBM to generate the class label candidates, we provide the top-$k$ classification accuracy of the CBMs we used, and this serves as the upper bound of our Chat-CBM when no intervention is conducted. The results are presented in Table~\ref{table:topk_classification_accuracy}. We can see that though the concept prediction of CBM is not perfect and the large language model is not fine-tuned on the target tasks, our Chat-CBM can still approach the theoretical upper limit of performance, which demonstrates the potential of our method.
\begin{table}[htb]
\centering
\caption{Top-N classification accuracy of CBMs on datasets with concept label. (ResNet-101 as backbone)}
\label{table:topk_classification_accuracy}
\resizebox{\linewidth}{!}{
\begin{tabular}{c|ccc|ccc|cccc}
\toprule
\textbf{Dataset} & \multicolumn{3}{c|}{\textbf{CUB}} & \multicolumn{3}{c|}{\textbf{AwA2}} & \multicolumn{4}{c}{\textbf{WBC}}\\
Top-N & 1 & 2 & 3 & 1 & 2 & 3 & 1 & 2 & 3 & 4 \\
\midrule
CBM & 0.752 & 0.825 & 0.848 & 0.923 & 0.969 & 0.978 & 0.988 & 0.992 & 0.998 & 1.000\\
Chat-CBM (LLaMA3-70B-Instruct) & \multicolumn{3}{c|}{0.815 (N2-K2)} & \multicolumn{3}{c|}{0.964 (N2-K2)} & \multicolumn{4}{c}{0.986 (N5-K3)}\\
\bottomrule
\end{tabular}
}
\end{table}

\subsection{Class Concept Semantics Prior Used for Different Datasets}\label{prior_knowledge_appendix}
Instead of using information from multiple places, such as Wikipedia, professional books, or websites, as prior knowledge. In the main experiments, we simply use the average concept of the class as the class concept semantics prior $\boldsymbol{\theta}$. That is, we statistically calculate the probabilities of different concepts appearing in the current class based on the concepts and class labels in the training set, and construct the prior knowledge accordingly. The prior knowledge used for different datasets is detailed below.

\begin{itemize}[noitemsep, topsep=0pt, leftmargin=*] % noitemsep,itemsep=1pt
\item \textbf{CUB}: We directly utilize the average concept label for each class and select concepts with an occurrence probability greater than 0.5. For example, the prior knowledge for the black-footed albatross class includes: ``bill shape is hooked seabird, underparts color is grey, breast pattern is solid, eye color is black, bill length is about the same as head, size is medium (9 - 16 in), back pattern is solid, tail pattern is solid, belly pattern is solid".

\item \textbf{AwA2}: The concept labels for AwA2 are originally class-level; we directly use these as the class prior knowledge. For example, ``antelope is usually associated with concepts including: furry, toughskin, big, lean, hooves, longleg, tail, chewteeth, horns, walks, fast, strong, muscle, quadrapedal, active, agility, vegetation, forager, grazer, newworld, oldworld, plains, fields, mountains, ground, timid, group".

\item \textbf{PBC}: We calculate the occurrence probability of each concept within each concept group for a given class in the training set. This is then used as the class prior knowledge. The specific representation is as follows: ``for Lymphocyte: cell\_size are mostly small, cell\_shape is mostly round, nucleus\_shape is mostly unsegmented-round, nuclear\_cytoplasmic\_ratio is high, chromatin\_density is densely, cytoplasm\_vacuole is no, cytoplasm\_texture is clear, cytoplasm\_color is light blue, granule\_type is nil, granule\_color is nil, granularity is no".

\item \textbf{Datasets without concept labels}: Given the absence of ground-truth concept labels, we identify the 10 most frequently occurring concepts for each class based on the validation set images. These are then used as the class prior knowledge. Take Labo trained on the Flower-102 dataset as an example: ``globe thistle is usually associated with concepts including: flower is also known as the blue thistle, thistle-like flower, attract bees, butterflies, and other pollinators, large, spiky, thistle-like flower, shaped like a thistle, self-seed itself, anti-inflammatory and healing properties, flower is also known as the bull thistle, not particularly attractive to bees or other pollinators, thistle is also known as the scotch thistle and is the national flower".
\end{itemize}

\subsection{Prompt and Output Format}\label{appendix:prompt_and_output_format}
\begin{table}[htb]
\caption{Prompt format for integrating prior knowledge on different classification tasks.}\label{tab:prompt_and_output_format}
\centering
\resizebox{\linewidth}{!}{
\begin{tabular}{l|l}
\toprule
Dataset & Format \\
\midrule
CUB  & ``\{classname\} usually has: \{concepts\}" \\
AwA2 & ``\{classname\} is usually associated with concepts including: \{concepts\}" \\
PBC  & ``for \{classname\}: \{concepts[i]\} is mostly / usually / (n\%) ..." \\
Other datasets & ``\{classname\} is usually associated with concepts including: \{concepts\}" \\
\bottomrule
\end{tabular}}
\end{table}

\subsection{Details about Intervention with New Concepts}\label{appendix:wiki}
\paragraph{Intervention under Controllable Incomplete Concept Settings.} 
We train independent CBMs on subsets of concepts from the CUB and AwA2 datasets. Specifically, we use 56/70/84/98/112 (full) concepts for CUB and 45/55/65/75/85 (full) concepts for AwA2, following the same hyperparameters described in Section~\ref{appendix:training_details}. These CBMs then serve as baseline models for the corresponding Chat-CBMs. Their classification performance of Chat-CBMs with LLaMA3-8B-Instruct is reported in Figure~\ref{fig:intervention_using_new_concepts}, where the starting point of each colored line is indicated by a triangle.  

For experiments involving interventions with new concepts, we augment the concept space step by step. At each step, 14 new concepts are introduced for CUB (10 for AwA2), but only those overlapping with the ground-truth labels of the image are integrated. The resulting performance of Chat-CBMs is shown with circular points in Figure~\ref{fig:intervention_using_new_concepts}.  

\paragraph{Intervention using Wikipedia Descriptions.} 
For CUB and AwA2, we further collect class-level feature descriptions from Wikipedia and use them as additional concepts to intervene in Chat-CBMs. The intervention prompt is:  
\textit{``In addition, we also know that $<$descriptions$>$. Answer again by considering the previous message and the new information."}  
To avoid class-label leakage, we replace the class name with general terms such as ``the bird" (CUB) or ``the animal" (AwA2). Two examples of such interventions are provided below.

\begin{itemize}[noitemsep, topsep=0pt, leftmargin=*] % noitemsep,itemsep=1pt
\item \textbf{black footed albatross}: The bird is a small member of the albatross family (while still large compared to most other seabirds) that has almost all black plumage. Some adults show white under tail coverts, and all adults have white markings around the base of the beak and below the eye. As the birds age, they acquire more white at the base of the beak. Its beak and feet are also all dark. They have only one plumage. They measure 68 to 74 cm (27-29 in), have a wingspan of 190 to 220 cm (6.2-7.2 ft), and weigh 2.6 to 4.3 kg (5.7-9.5 lb). Males, at an average weight of 3.4 kg (7.5 lb), are larger than females, at an average of 3 kg (6.6 lb).

\item \textbf{beaver}: The animals are the second-largest living rodents. The animals have large skulls with powerful chewing muscles. They have four chisel-shaped incisors that continue to grow throughout their lives. The incisors are covered in a thick enamel that is colored orange or reddish-brown by iron compounds. The lower incisors have roots that are almost as long as the entire lower jaw. Animals have one premolar and three molars on all four sides of the jaws, adding up to 20 teeth. The molars have meandering ridges for grinding woody material. The eyes, ears, and nostrils are arranged so that they can remain above water while the rest of the body is submerged. The nostrils and ears have valves that close underwater, while nictitating membranes cover the eyes.
\end{itemize}

\subsection{The Metadata for Line Plot}
The numerical metadata of Figure~\ref{fig:intervention_on_dataset_with_concept_labels}, Figure~\ref{fig:intervention_unsupervised_datasets}, and Figure~\ref{fig:intervention_using_new_concepts} are detailed as follows.

\begin{table}[!h]
    \centering
    \caption{Metadata (classification accuracy) for the CUB dataset in Figure~\ref{fig:intervention_on_dataset_with_concept_labels}.}
    \resizebox{\linewidth}{!}{
    \begin{tabular}{l|cccccccccccc}
    \toprule
    \diagbox{model}{ratio} & 0.0 & 0.1 & 0.2 & 0.3 & 0.4 & 0.5 & 0.6 & 0.7 & 0.8 & 0.9 & 1.0 \\
    \midrule
    Hard CBM & 0.7076 & 0.7378 & 0.7672 & 0.7995 & 0.8533 & 0.8931 & 0.9308 & 0.9586 & 0.9838 & 0.9960 & 0.9983\\ 
    CBM      & 0.7525 & 0.7846 & 0.8112 & 0.8460 & 0.8749 & 0.8918 & 0.9134 & 0.9326 & 0.9513 & 0.9694 & 0.9738\\
    ECBM     & 0.8063 & 0.8133 & 0.8143 & 0.8194 & 0.8475 & 0.8702 & 0.9046 & 0.9263 & 0.9327 & 0.9549 & 0.9801\\
    CEM      & 0.7991 & 0.8261 & 0.8472 & 0.9121 & 0.9303 & 0.9420 & 0.9593 & 0.9661 & 0.9803 & 0.9855 & 0.9911\\
    Chat-CBM & 0.7978 & 0.8886 & 0.9384 & 0.9617 & 0.9778 & 0.9874 & 0.9921 & 0.9943 & 0.9967 & 0.9975 & 0.9984\\
    \bottomrule
    \end{tabular}}
\end{table}

\begin{table}[!h]
    \centering
    \caption{Metadata (classification accuracy) for the AwA2 dataset in Figure~\ref{fig:intervention_on_dataset_with_concept_labels}.}
    \resizebox{\linewidth}{!}{
    \begin{tabular}{l|ccccccccccccc}
    \toprule
    \diagbox{model}{ratio} & 0.00 & 0.09 & 0.18 & 0.27 & 0.36 & 0.45 & 0.55 & 0.64 & 0.73 & 0.82 & 0.91 & 1.00 \\
    \midrule
    Hard CBM & 0.9006 & 0.9351 & 0.9526 & 0.9575 & 0.9661 & 0.9721 & 0.9766 & 0.9854 & 0.9914 & 0.9932 & 0.9950 & 0.9950\\ 
    CBM      & 0.9233 & 0.9300 & 0.9475 & 0.9635 & 0.9728 & 0.9786 & 0.9826 & 0.9894 & 0.9939 & 0.9950 & 0.9974 & 0.9974\\
    ECBM     & 0.9160 & 0.9217 & 0.9266 & 0.9375 & 0.9491 & 0.9773 & 0.9864 & 0.9932 & 0.9951 & 0.9961 & 0.9973 & 0.9973\\
    CEM      & 0.9242 & 0.9411 & 0.9583 & 0.9727 & 0.9831 & 0.9947 & 0.9982 & 0.9989 & 0.9992 & 0.9995 & 0.9996 & 0.9996\\
    Chat-CBM & 0.9572 & 0.9732 & 0.9805 & 0.9856 & 0.9881 & 0.9930 & 0.9947 & 0.9954 & 0.9962 & 0.9989 & 0.9996 & 0.9996\\
    \bottomrule
    \end{tabular}}
\end{table}

\begin{table}[!h]
    \centering
    \caption{Metadata (classification accuracy) for the PBC dataset in Figure~\ref{fig:intervention_on_dataset_with_concept_labels}.}
    \resizebox{\linewidth}{!}{
    \begin{tabular}{l|cccccccccccccccc}
    \toprule
    \diagbox{model}{ratio} & 0.00 & 0.07 & 0.14 & 0.21 & 0.29 & 0.36 & 0.43 & 0.50 & 0.57 & 0.64 & 0.71 & 0.79 & 0.86 & 0.93 & 1.00 \\
    \midrule
    Hard CBM & 0.9593 & 0.9606 & 0.9678 & 0.9681 & 0.9684 & 0.9808 & 0.9812 & 0.9804 & 0.9806 & 0.9810 & 0.9823 & 0.9854 & 0.9871 & 0.9875 & 0.9883\\ 
    CBM      & 0.9880 & 0.9837 & 0.9834 & 0.9804 & 0.9804 & 0.9831 & 0.9819 & 0.9782 & 0.9780 & 0.9781 & 0.9791 & 0.9802 & 0.9805 & 0.9802 & 0.9799\\
    ECBM     & 0.9941 & 0.9931 & 0.9944 & 0.9945 & 0.9945 & 0.9955 & 0.9962 & 0.9962 & 0.9966 & 0.9973 & 0.9981 & 0.9989 & 0.9991 & 0.9993 & 0.9994\\
    CEM      & 0.9933 & 0.9943 & 0.9955 & 0.9967 & 0.9967 & 0.9977 & 0.9975 & 0.9974 & 0.9981 & 0.9983 & 0.9985 & 0.9987 & 0.9995 & 0.9996 & 0.9998\\
    Chat-CBM & 0.9764 & 0.9803 & 0.9822 & 0.9911 & 0.9928 & 0.9941 & 0.9956 & 0.9966 & 0.9971 & 0.9972 & 0.9979 & 0.9983 & 0.9988 & 0.9981 & 0.9989\\
    \bottomrule
    \end{tabular}}
\end{table}

\begin{table}[!h]
    \centering
    \caption{Metadata (classification accuracy) for Figure~\ref{fig:intervention_unsupervised_datasets}. N. denotes the number of interventions.}
    \resizebox{\linewidth}{!}{
    \begin{tabular}{l|cccccc|cccccc}
    \toprule
    \multirow{2}*{\diagbox{Dataset}{N.}} & \multicolumn{6}{c|}{Chat-CBM (V2C-CBM)} & \multicolumn{6}{c}{Chat-CBM (LaBo)} \\
    & 0 & 1 & 2 & 3 & 4 & 5 & 0 & 1 & 2 & 3 & 4 & 5 \\
    \midrule
    CIFAR10         & 0.955 & 0.977 & 0.992 & 0.997 & 0.999 & 0.999 & 0.889 & 0.912 & 0.955 & 0.967 & 0.976 & 0.981\\
    CIFAR100        & 0.727 & 0.813 & 0.881 & 0.895 & 0.935 & 0.965 & 0.670 & 0.766 & 0.815 & 0.848 & 0.899 & 0.932\\
    Food-101        & 0.786 & 0.855 & 0.916 & 0.937 & 0.963 & 0.979 & 0.753 & 0.837 & 0.897 & 0.925 & 0.949 & 0.964\\
    Flower-102      & 0.914 & 0.946 & 0.978 & 0.988 & 0.993 & 0.997 & 0.876 & 0.956 & 0.980 & 0.990 & 0.995 & 0.997\\
    DTD             & 0.734 & 0.868 & 0.926 & 0.952 & 0.965 & 0.977 & 0.677 & 0.875 & 0.955 & 0.978 & 0.984 & 0.988 \\
    \bottomrule
    \end{tabular}}
\end{table}

\begin{table}[!ht]
    \centering
    \caption{Metadata for Figure~\ref{fig:intervention_using_new_concepts}. N. denotes the number of concepts. We use start=1,2,3,4,5 to represent the corresponding starting number of concepts for the CUB and AwA2 datasets.}
    \resizebox{\linewidth}{!}{
    \begin{tabular}{l|cccccc|cccccc}
    \toprule
    \multirow{2}*{\diagbox{start=}{N.}} & \multicolumn{6}{c|}{CUB} & \multicolumn{6}{c}{AwA2} \\
      & 56 & 70 & 84 & 98 & 112 & +wiki & 45 & 55 & 65 & 75 & 85 & +wiki \\
     \midrule
     1 & 0.620 & 0.685 & 0.729 & 0.769 & 0.805 & - & 0.901 & 0.924 & 0.943 & 0.950 & 0.959 & -\\
     2 & - & 0.675 & 0.736 & 0.798 & 0.858 & - & - & 0.920 & 0.945 & 0.953 & 0.960 & -\\
     3 & - & - & 0.727 & 0.817 & 0.893 & - & - & - & 0.934 & 0.954 & 0.961 & -\\
     4 & - & - & - & 0.760 & 0.912 & - & - & - & - & 0.949 & 0.967 & - \\
     5 & - & - & - & - & 0.797 & 0.945 & - & - & - & - & 0.957 & 0.989\\
     \bottomrule
    \end{tabular}
    }
    \label{tab:placeholder}
\end{table}

\section{Limitations and Future Work}\label{appendix:limitation}
\paragraph{Hard to learning additional information under incomplete concept supervision.}
Chat-CBMs leverage the concept semantics for reasoning, so they inherently prevent possible concept leakage problems of existing CBMs, in which some concept activations may actually serve as a class label proxy. But the structure of Chat-CBM also makes it challenging to learn additional information to improve performance when concept label discriminability is very insufficient, as explored in Concept Embedding Models (CEM)~\citep{cem_neurips2022}. 

However, in cases where only a portion of the concept set is incomplete (which we think is a more common case), Chat-CBM's strong generalization and few-shot capabilities will still allow its performance to be comparable to or better than baselines. To validate this, we train independent CBMs using reduced subsets of the original concept labels (56/70/84/98/112 for CUB; 45/55/65/75/85 for AwA2), and then evaluate them. The results in Table~\ref{tab:concept_incomplete_cub} and Table~\ref{tab:concept_incomplete_awa2} validate that Chat-CBMs still achieve better performance compared to CBMs under this setting.
\begin{table}[htbp]
    \centering
    \caption{Performance of CBMs and Chat-CBMs under incomplete concepts on the CUB dataset.}
    \begin{tabular}{l|ccccc}
    \toprule
    Number of Concepts & 56 & 70 & 84 & 98 & 112\\
    \midrule
    Independent CBMs                      & 0.591 & 0.666 & 0.712 & 0.743 & 0.752 \\
    Chat-CBMs (LLaMA3-8B-Instruct)        & 0.620 & 0.675 & 0.727 & 0.760 & 0.797 \\
    \bottomrule
    \end{tabular}
    
    \label{tab:concept_incomplete_cub}
\end{table}

\begin{table}[htbp]
    \centering
    \caption{Performance of CBMs and Chat-CBMs under incomplete concepts on the AwA2 dataset.}
    \begin{tabular}{l|ccccc}
    \toprule
    Number of Concepts  & 45 & 55 & 65 & 75 & 85\\
    \midrule
    Independent CBMs                        & 0.913 & 0.915 & 0.921 & 0.922 & 0.923 \\
    Chat-CBMs (LLaMA3-8B-Instruct)          & 0.901 & 0.920 & 0.934 & 0.949 & 0.957 \\
    \bottomrule
    \end{tabular}
    
    \label{tab:concept_incomplete_awa2}
\end{table}

\paragraph{Computational Cost.}
Using an LLM as a language-based classifier typically means an extra 10+~100+ GBs of GPU memory per image (depending on the LLM size), and an average of 3.32 sec/image for generating complete outputs (until the EOS token) on an NVIDIA L40 GPU using LLaMA-3-8B-Instruct. While this does increase computational cost and latency for large-scale experiments (e.g., testing Chat-CBM performance on new benchmarks), it's acceptable for user-facing interactive use cases. The streaming output style, the widespread availability, and the rapid response of LLM APIs can further mitigate this influence. For large-scale deployment, given that we retain the explicit concept bottleneck structure in Chat-CBMs, we can also cache common concept input contexts during the model's actual service phase, thereby reducing operational costs and improving inference speed. However, there is no denying that Chat-CBM does introduce significant computational overhead compared to the standard CBM architecture.

\paragraph{Potential Harmful Knowledge in LLMs.}
As Chat-CBM builds on frozen LLMs as label predictors, it inevitably inherits the knowledge embedded in these models. While this enables strong semantic reasoning, it also carries potential risks: LLMs may encode harmful, biased, or misleading knowledge, which could in turn affect the model’s predictions or the interaction process. Although our experiments are limited to benchmark datasets and do not involve deployment in high-stakes applications, these risks must be considered before applying Chat-CBM in sensitive domains such as medicine or law. Future work should incorporate alignment strategies (e.g., safety-tuned LLMs~\citep{wu2025mitigating}, safeguarding~\citep{wei2023jailbreak,wei2025rega}, or unlearning the harmful knowledge~\citep{wu2025reliable}) to ensure that user-facing interventions remain safe, unbiased, and reliable.

\section{More Visualization Results}\label{section::more_vis}
\begin{figure}[htbp]
    \centering
    \includegraphics[width=\linewidth]{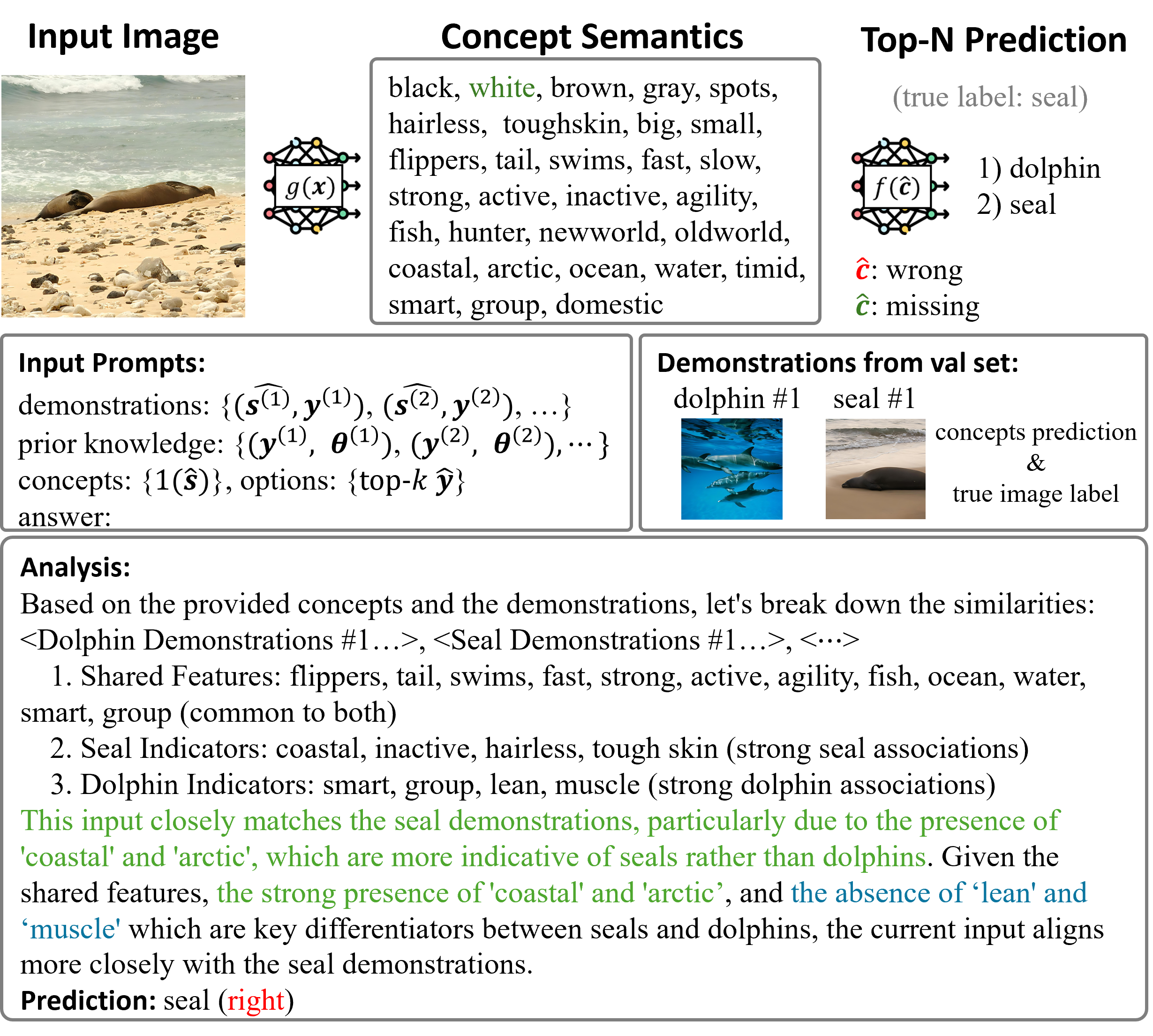}
    \caption{Visualization of the inference process of Chat-CBM on the AwA2 dataset. \textcolor{green_color}{Green} highlights the positive reasoning and \textcolor{blue_color}{blue} highlights the negative reasoning process.}
    \label{fig:visualization_awa2_common_1}
\end{figure}

\begin{figure}[htbp]
    \centering
    \includegraphics[width=\linewidth]{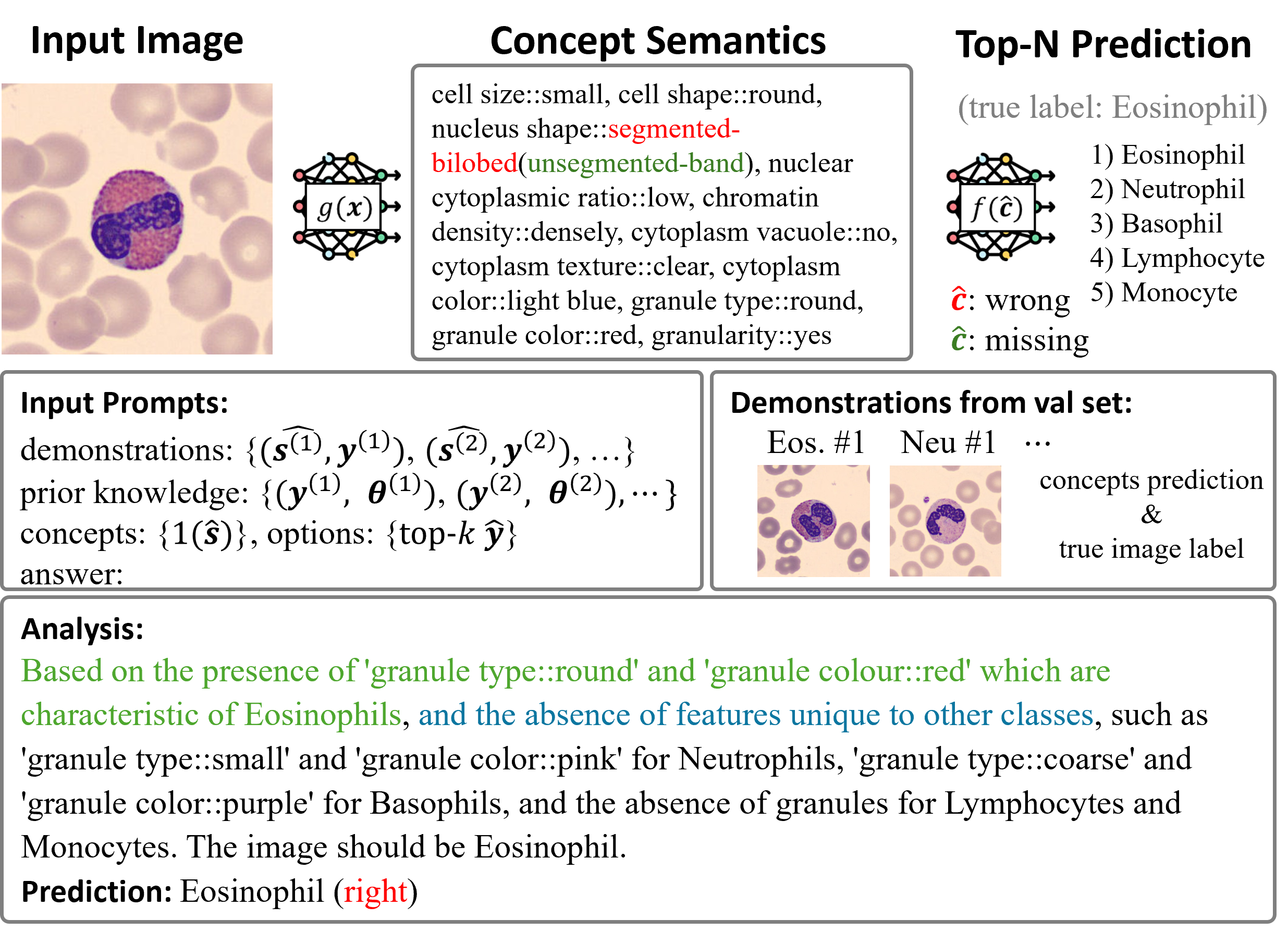}
    \caption{Visualization of the inference process of Chat-CBM on the PBC dataset. \textcolor{green_color}{Green} highlights the positive reasoning and \textcolor{blue_color}{blue} highlights the negative reasoning process.}
    \label{fig:visualization_pbc_common_1}
\end{figure}

\begin{figure}[htbp]
    \centering
    \includegraphics[width=\linewidth]{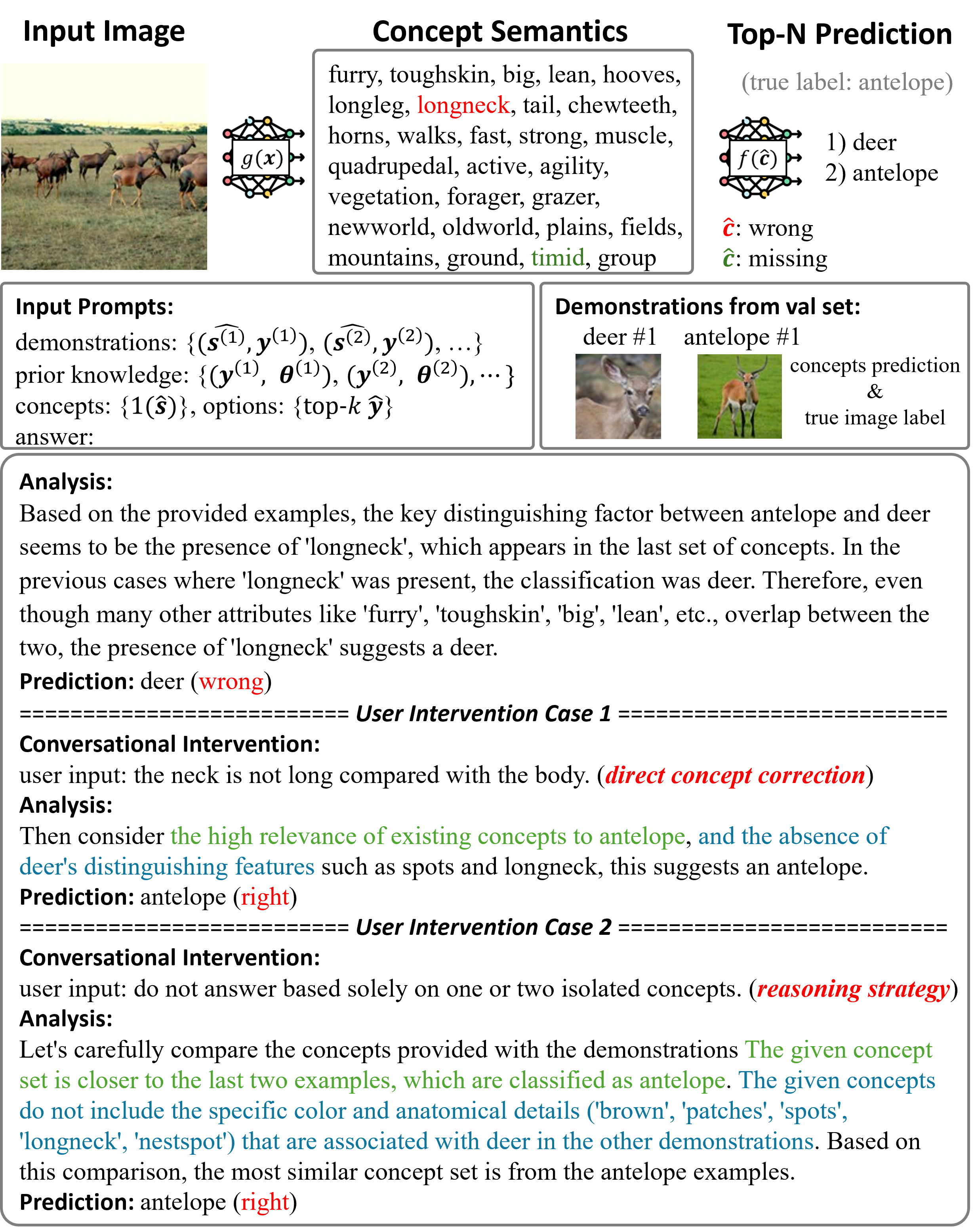}
    \caption{Visualization of the conversational intervention process of Chat-CBM on the AwA2 dataset. \textcolor{green_color}{Green} highlights the positive reasoning and \textcolor{blue_color}{blue} highlights the negative reasoning process. Users can either directly correct concept predictions like standard CBMs (\textbf{case 1}) or give a high-level reasoning strategy to guide thinking (\textbf{case 2}).}
    \label{fig:visualization_awa2_intervention_1}
\end{figure}

\begin{figure}[htbp]
    \centering
    \includegraphics[width=\linewidth]{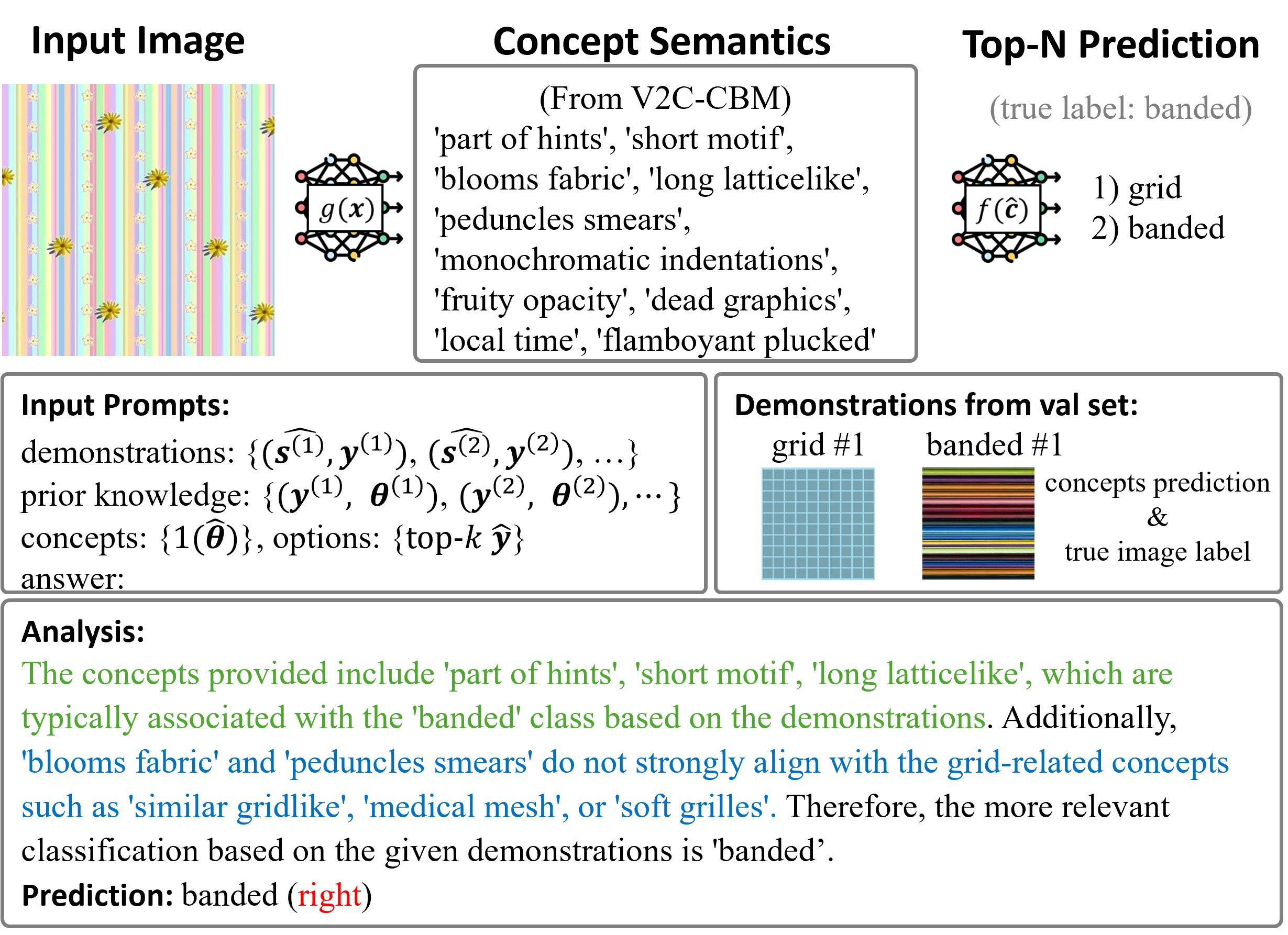}
    \caption{Visualization of the inference process of Chat-CBM on the DTD dataset. The concept bank is from V2C-CBM.}
    \label{fig:visualization_DTD_common_1}
\end{figure}

\begin{figure}[htbp]
    \centering
    \includegraphics[width=\linewidth]{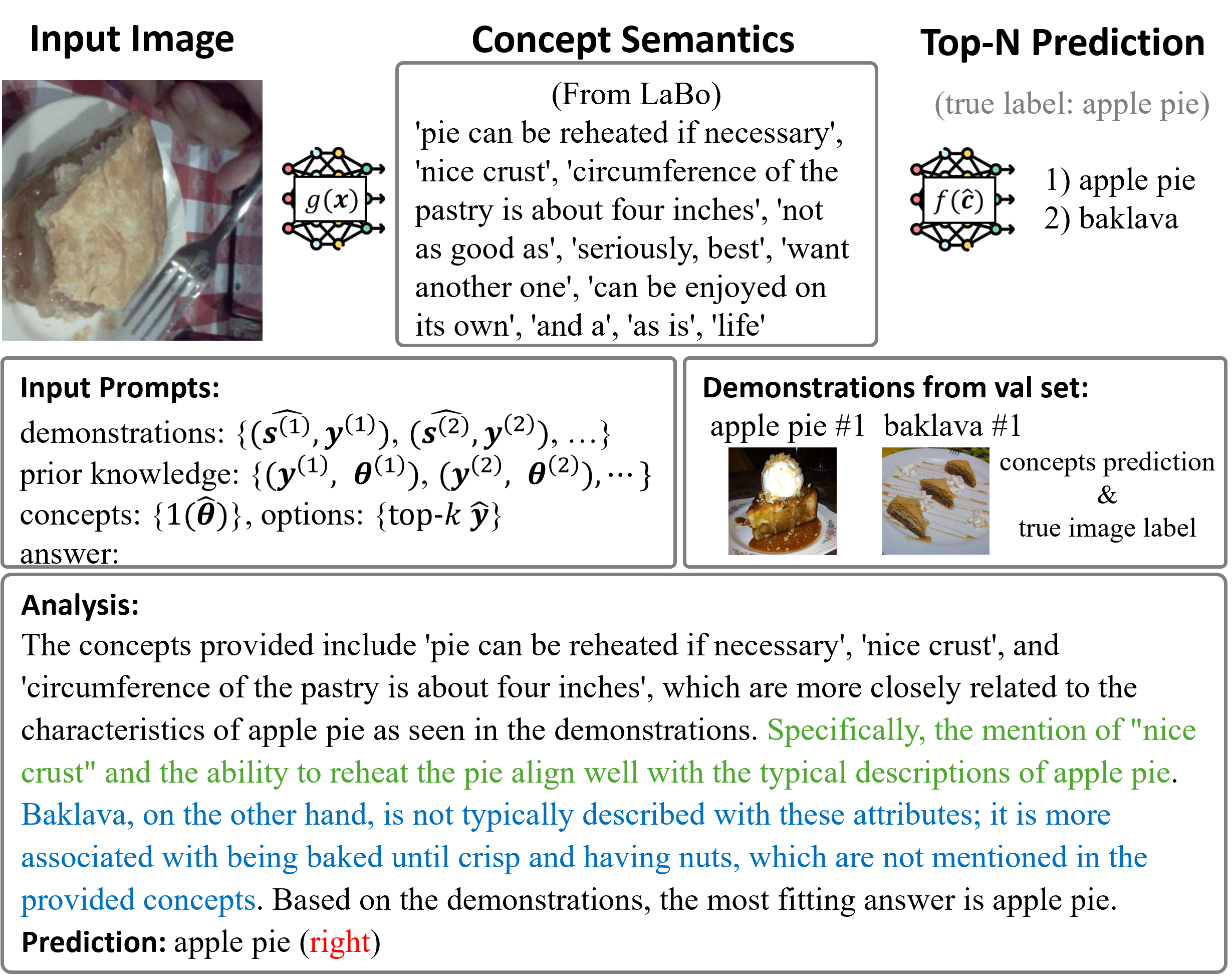}
    \caption{Visualization of the inference process of Chat-CBM on the Food-101 dataset. The concept bank is from LaBo.}
    \label{fig:visualization_food_common_1}
\end{figure}

We provide more examples of the reasoning process or the intervention process of Chat-CBMs in Figure~\ref{fig:visualization_awa2_common_1},~\ref{fig:visualization_pbc_common_1}, and~\ref{fig:visualization_awa2_intervention_1} for datasets with concept labels. And Figure~\ref{fig:visualization_DTD_common_1} and Figure~\ref{fig:visualization_food_common_1} are for datasets without concept labels.

\end{document}

%% file: math_commands.tex
\usepackage{amsmath,amsfonts,bm}

\def\eqref#1{equation~\ref{#1}}
\def\1{\bm{1}}

\DeclareMathAlphabet{\mathsfit}{\encodingdefault}{\sfdefault}{m}{sl}
\SetMathAlphabet{\mathsfit}{bold}{\encodingdefault}{\sfdefault}{bx}{n}